%% file: 0_main.tex
\documentclass[final,1p,times]{elsarticle}
\usepackage{amssymb}
\usepackage{amsmath}
\usepackage{booktabs}
\usepackage{multirow}

\usepackage{tabularx} % 放到导言区
\usepackage{arydshln}
\usepackage{hyperref}
\usepackage{natbib}

\journal{Journal of Biomedical Informatics}

\begin{document}

\begin{frontmatter}



\title{Retrieval-augmented in-context learning for multimodal large language models in disease classification} %% Article title

\author[label1]{Zaifu Zhan}
\author[label2]{Shuang Zhou}
\author[label3]{Xiaoshan Zhou}
\author[label4]{Yongkang Xiao}
% \author[label4]{Han Yang}
\author[label2]{Jun Wang}
\author[label5]{Jiawen Deng}
\author[label6]{He Zhu}
% \author[label4]{Huixue Zhou}
\author[label2]{Yu Hou}
% \author[label7]{Mingchen Li}
\author[label2]{Rui Zhang\corref{cor1}}

\affiliation[label1]{
organization={Department of Electrical and Computer Engineering, University of Minnesota},
addressline={200 Union St SE},
city={Minneapolis},
postcode={55455},
state={MN},
country={USA}}
\affiliation[label2]{
organization={Division of Computational Health Sciences, Department of Surgery, University of Minnesota},
addressline={516 Delaware St SE},
city={Minneapolis},
postcode={55455},
state={MN},
country={USA}}
\affiliation[label3]{
organization={Department of Civil and Environmental Engineering, University of Michigan},
addressline={2350 Hayward Street},
city={Ann Arbor},
postcode={48109},
state={MI},
country={USA}
}
\affiliation[label4]{
organization={Institute for Health Informatics, University of Minnesota},
addressline={516 Delaware Street SE},
city={Minneapolis},
postcode={55455},
state={MN},
country={USA}}
\affiliation[label5]{
organization={Department of Computer Science and Engineering, University of Minnesota},
addressline={200 Union St SE},
city={Minneapolis},
postcode={55455},
state={MN},
country={USA}}
\affiliation[label6]{
organization={Department of Chemical Engineering and Materials Science, University of Minnesota},
addressline={421 Washington Ave. SE, Minneapolis},
city={Minneapolis},
postcode={55455},
state={MN},
country={USA}}
% \affiliation[label7]{
% organization={Manning College of Information and Computer Sciences, University of Massachusetts Amherst},
% addressline={300 Massachusetts Ave},
% city={Amherst},
% postcode={01003},
% state={MA},
% country={USA}}

\cortext[cor1]{Corresponding author: Rui Zhang (email: ruizhang@umn.edu)}

% \affiliation[label2]{organization={},

% \author{} %% Author name

% \affiliation{organization={},%Department and Organization

%% Abstract
\begin{abstract}
\textbf{Objectives}:
We aim to dynamically retrieve informative demonstrations, enhancing in-context learning in multimodal large language models (MLLMs) for disease classification.

\noindent
\textbf{Methods}:
We propose a Retrieval-Augmented In-Context Learning (RAICL) framework, which integrates retrieval-augmented generation (RAG) and in-context learning (ICL) to adaptively select demonstrations with similar disease patterns, enabling more effective ICL in MLLMs.
Specifically, RAICL examines embeddings from diverse encoders, including ResNet, BERT, BioBERT, and ClinicalBERT, to retrieve appropriate demonstrations, and constructs conversational prompts optimized for ICL. 
We evaluated the framework on two real-world multi-modal datasets (TCGA and IU Chest X-ray), assessing its performance across multiple MLLMs (Qwen, Llava, Gemma), embedding strategies, similarity metrics, and varying numbers of demonstrations.

\noindent
\textbf{Results}:
RAICL consistently improved classification performance. Accuracy increased from 0.7854 to 0.8368 on TCGA and from 0.7924 to 0.8658 on IU Chest X-ray. Multi-modal inputs outperformed single-modal ones, with text-only inputs being stronger than images alone. 
The richness of information embedded in each modality will determine which embedding model can be used to get better results.
Few-shot experiments showed that increasing the number of retrieved examples further enhanced performance. Across different similarity metrics, Euclidean distance achieved the highest accuracy while cosine similarity yielded better macro-F1 scores. RAICL demonstrated consistent improvements across various MLLMs, confirming its robustness and versatility.

\noindent
\textbf{Conclusions}:
RAICL provides an efficient and scalable approach to enhance in-context learning in MLLMs for multimodal disease classification.
\end{abstract}


% \begin{highlights}
% \end{highlights}

%% Keywords
\begin{keyword}
Retrieval-augmented generation \sep Multimodal large language models \sep In-context learning \sep Disease classification.
\end{keyword}

\end{frontmatter}



\input{1_Introduction}
\input{2_methods}

\input{3_results}
\input{4_Discussion}
\input{5_conclusion}

\bibliographystyle{elsarticle-num} 
\bibliography{0_main}

\end{document}

%% file: 1_Introduction.tex
\section{Introduction}
\label{sec_intro}

Multi-modality disease classification leverages various sources of patient information, such as medical images and clinical notes, to infer the most likely diagnosis from a set of related diseases, offering significant clinical value~\cite{yu2023multi,zhang2023multi,liu2025progressive}. Integrating multiple modalities of clinical data allows for a comprehensive depiction of a patient’s health status, thereby enhancing diagnostic accuracy and reliability~\cite{han2024comparative}.
For instance, medical images, including histopathology slides and radiographs, provide detailed visual insights into the development of pathological cells or structural abnormalities~\cite{gurcan2009histopathological, drew2013informatics}. Meanwhile, clinical reports and pathology notes deliver complementary textual information, such as patient history, symptoms, and contextual details, that can help interpret visual findings and uncover subtle patterns that might otherwise be missed~\cite{nissen2014clinical, adhikari2024multi}. 
The synergistic integration of these modalities ensures a more holistic understanding of a patient’s condition~\cite{xu2024comprehensive,niu2024text}.

Various multi-modality disease classification models have been proposed, achieving remarkable performance, particularly in supervised learning settings~\cite{yu2023multi,moglia2024minigpt,kumar2025transformer}. Early studies primarily relied on extensive labeled data to train models capable of effectively integrating multi-modal inputs. For example, CheXpert~\cite{irvin2019chexpert} utilized over 224,316 labeled chest X-rays paired with textual radiology reports to train deep learning models for multi-label thoracic disease classification, achieving an AUROC of 0.93 for pneumonia classification. Another notable study, MedFuse~\cite{hayat2022medfuse}, integrated multi-modal information from over 10,000 annotated samples of pathological images and clinical notes to enhance diagnostic accuracy in skin cancer classification, achieving an improvement of 8\% in accuracy compared to image-only models. Despite these successes, extensive (e.g., a few hundred thousand) well-annotated multi-modal samples, such as paired image-text data, are often unavailable in practice. This limitation underscores the need for alternative approaches to leverage less labeled data while maintaining high performance.

Recent advancements in multimodal large language models (MLLMs)~\cite{zhang2024mm,zhou2024large}, such as GPT-4o~\cite{achiam2023gpt}, DeepSeek~\cite{guo2025deepseek}, and Qwen~\cite{bai2025qwen2}, have demonstrated significant potential in addressing the above issue with two key advantages. 
First, MLLMs excel at integrating both visual and textual information from patients’ clinical data, facilitating more comprehensive clinical decision-making~\cite{niu2024text}. Second, these models support prompting methods~\cite{wu2024leveraging,chakraborty2025prompt,tai2024link} that require only a few demonstration examples for in-context learning, thereby reducing the reliance on large amounts of labeled data while maintaining remarkable performance~\cite{zhan2025mmrag}. Given that small amounts of labeled data are often available in clinical settings~\cite{pai2024foundation}, the use of MLLMs for multi-modal disease classification has garnered growing interest. 
For example, Ferber et al.\cite{ferber2024context} demonstrated that in-context learning enables GPT-4V to match or even surpass the performance of specialized neural networks trained for specific tasks, while requiring only a minimal number of examples. Their evaluation focused on three critical cancer histopathology tasks: classification of tissue subtypes in colorectal cancer, colon polyp subtyping, and breast tumor detection in lymph node sections.
Similarly, Jiang et al.\cite{jiang2024many} benchmarked GPT-4o and Gemini 1.5 Pro on 10 datasets covering a range of tasks—including multi-class, multi-label, and fine-grained disease classification—and found that providing more examples in the prompt consistently led to better performance than using fewer examples.

Despite these advances, the current efforts suffer from a critical limitation, which renders sub-optimal performance on this task. Specifically, the in-context length of MLLMs is typically constrained~\cite{li2024long,song2024milebench}, whereas real-world patient data, such as clinical notes, are often lengthy. This disparity makes it difficult to include all relevant demonstrations in the prompt. Given this constraint, existing studies generally randomly select a subset of demonstration examples for in-context learning~\cite{jiang2024many,li2024improving}. However, the choice of demonstrations can significantly influence the MLLMs' performance in clinical decision-making~\cite{liu2025application,oniani2024enhancing}. 
Intuitively, demonstration samples that share similar disease profiles, such as matching signs and symptoms in clinical notes or similar pathological cell morphologies, are likely to serve as more informative references and positively influence the model's predictions. 
This insight has been supported by related research~\cite{peng2024revisiting, xu2024introspection}. 
% For example, xxx reported that selecting demonstrations with similar clinical attributes, such as patient demographics and disease history, enhanced the accuracy of xxx tasks from xxx to xxx.
For example, Peng et al.~\cite{peng2024revisiting} showed that the performance of a demonstration positively correlates with its contribution to the model’s understanding of the test samples.
And Xu et al.~\cite{xu2024introspection}
concluded that demonstrations selected according to textual similarity improve the performance of MLLMs.
Therefore, identifying and selecting ``appropriate'' demonstrations from the available labeled data holds promise for improving the effectiveness of prompting MLLMs in multi-modal disease classification. However, this challenge remains largely under-explored and warrants further investigation.

To bridge the gap, we proposed a Retrieval-Augmented In-Context Learning (RAICL) framework, which dynamically retrieved informative examples tailored to the test sample as demonstrations for effective in-context learning. The core idea lies in leveraging embedding similarity as a proxy to identify informative references, where samples with high semantic similarity are assumed to present similar disease patterns. 
Specifically, RAICL systematically examined the widely used encoders, including ResNet, BERT, BioBERT, and ClinicalBERT, to generate semantic embeddings for image and textual data, respectively. It further investigated diverse similarity metrics for embedding similarity measurements. Extensive experiments across various MLLMs like Qwen~\cite{bai2025qwen2}, Llava~\cite{liu2024llavanext}, and Gemma~\cite{team2025gemma} on two real-world datasets, i.e., TCGA~\cite{weinstein2013cancer} and the IU Chest X-ray~\cite{demner2016preparing}, verified the effectiveness of RAICL for multi-modal disease classification. Notably, the framework demonstrates robust performance under few-shot settings, highlighting its adaptability and resilience in scenarios with limited labeled data. Overall, we summarize the significance statement below.

% To improve the performance of MLLMs in disease classification, we proposed a retrieval-augmented in-context learning framework. This noval framework adopted the retrieval-augmented generation (RAG)~\cite{lewis2020retrieval,zhan2025mmrag,zhan2025ramie,li2024benchmarking} and in-context learning (ICL)~\cite{zhan2025mmrag,liu2024context,gutierrez2022thinking} techniques.
% ICL provides a promising training-free method for MLLMs to achieve high performance. 
% Specifically, the ICL technique adapts models to new tasks by leveraging a set of relevant examples during inference, further enhancing the capabilities of MLLMs.
% At the same time, RAG, which was designed to retrieve related information to reduce hallucination~\cite{bai2024hallucination,ye2023cognitive}, now fills in the framework to fetch the most relevant examples for ICL.
% In this paper, we demonstrate the effectiveness of our framework on two publicly available yet real-world datasets. Through a series of experiments, we show that our framework is highly versatile across different MLLMs and can enhance performance in few-shot settings. Finally, we prove its compatibility with various similarity metrics, highlighting its exceptional robustness. Overall, the statement of significance of our work is in table \ref{tab:sos}.

\begin{table}[h]
    \centering
    \begin{tabularx}{\textwidth}{p{4cm}|X}
    \toprule
    Statement of Significance.\\
    \midrule
    Problem & The selection of appropriate demonstrations for MLLMs' in-context learning in multi-modality disease classification remains an underexplored area of research.\\
    \midrule
    What is already known & The choice of demonstration examples in the prompt may influence the performance of MLLMs. \\
    \midrule
    What this paper adds & We proposed a novel RAICL framework that integrates RAG and ICL with customized conversational prompts to retrieve informative demonstrations and effectively guide MLLMs in disease classification.\\
    \midrule
    Who would benefit from the new knowledge in this paper & Researchers working with MLLMs to facilitate clinical decision-making.\\
    \bottomrule
    \end{tabularx}
    \label{tab:sos}
\end{table}

%% file: 2_methods.tex
\section{Method}
\label{sec_med}

\subsection{Task Formulation}

A data sample is represented as \(s = (x, r)\), where  
\(x \in \mathbb{R}^{H \times W \times D}\) denotes a generic medical image tensor, with \(H\), \(W\), and \(D\) representing the height, width, and number of channels, respectively.  
\(r\) is an accompanying free-text description providing additional context or clinical notes.

Given a dataset $\mathcal{D} = \{(x_i, r_i, y_i)\}_{i=1}^{N}$,
where \(x_i\) is the \(i\)-th image, \(r_i\) is the corresponding text description, and \(y_i\) is the ground-truth label (e.g., a classification result),  
the objective for a multi-modality large language model \(G\) parameterized by \(\theta\) is to generate a label string  
\[
\hat{y} = f(G_\theta(x, r)),
\]  
where \(G_\theta(x, r)\) denotes the model output given both the image and the text as input, and \(f(\cdot)\) is a decoding or post-processing function that maps the model's raw output to a discrete label string.  
The predicted label \(\hat{y}\) is expected to exactly match the ground-truth label \(y\).

\subsection{Datasets}
The first dataset used is a subset of The Cancer Genome Atlas (TCGA) collections~\cite{weinstein2013cancer}, specifically breast cancer (BRCA), uterine corpus endometrial carcinoma (UCEC), lower grade glioma (LGG), lung adenocarcinoma (LUAD), and bladder cancer (BLCA) projects.  
This dataset consists of high-resolution histopathology whole-slide images (WSIs) collected from multiple institutions, each annotated with detailed cancer type diagnoses.  
The WSIs, scanned at magnifications typically ranging from 20$\times$ to 40$\times$, capture fine-grained tissue morphology crucial for disease characterization.  
TCGA provides a valuable resource for multimodal large language models in disease classification, as the histological complexity and inter-patient variability challenge the model's ability to extract and reason over visual patterns.  
Moreover, the diversity of cancer types enables evaluation of the model's generalization capacity across different pathological domains.

The second dataset is the Indiana University Chest X-ray (IU-CXR) collection~\cite{demner2016preparing}, containing 7,470 chest radiographs paired with 3,955 corresponding radiology reports.  
The dataset includes both frontal and lateral views, covering a wide range of thoracic conditions documented in structured textual findings and impressions.  
Each image-report pair has been annotated into several common disease categories, allowing for a multi-class prediction setting.  
IU-CXR is highly suitable for MLLM-based disease classification tasks, as it naturally combines free-text clinical descriptions with medical imaging, requiring the model to integrate heterogeneous modalities.  
Furthermore, the dataset reflects real-world diagnostic variability and uncertainty, offering a realistic benchmark to assess the robustness of multimodal generation in clinical scenarios. 
We filtered out data that omitted images, text, or had multiple diseases.

\begin{table}[ht]
\centering
\caption{Summary of datasets used in this study.}
\label{tab:dataset_summary}
\begin{tabular}{lc}
\toprule
\textbf{Class} & \textbf{Number of Samples} \\
\midrule
\multicolumn{2}{c}{\textbf{TCGA Dataset}} \\
\midrule
BRCA & 900 \\
UCEC & 479 \\
LGG & 435 \\
LUAD & 429 \\
BLCA & 342 \\
\midrule
\multicolumn{2}{c}{\textbf{IU Chest X-ray Dataset}} \\
\midrule
 Normal & 1197 \\
 Calcified Granuloma & 72 \\
 Calcinosis & 33 \\
 Opacity & 28 \\
 Scoliosis & 24 \\
 Cardiomegaly & 24 \\ 
 Spondylosis & 20 \\
 Granulomatous Disease & 18 \\
 Fractures & 16 \\
 Osteophyte & 15 \\
 Nodule & 11 \\
 Foreign Bodies & 11 \\
 Atherosclerosis & 10 \\
 Cicatrix & 9 \\
 Granuloma & 5 \\
 Pulmonary Atelectasis & 5 \\
\bottomrule
\end{tabular}
\end{table}

\begin{figure}[htbp]
    \centering
    \includegraphics[width=0.9\linewidth]{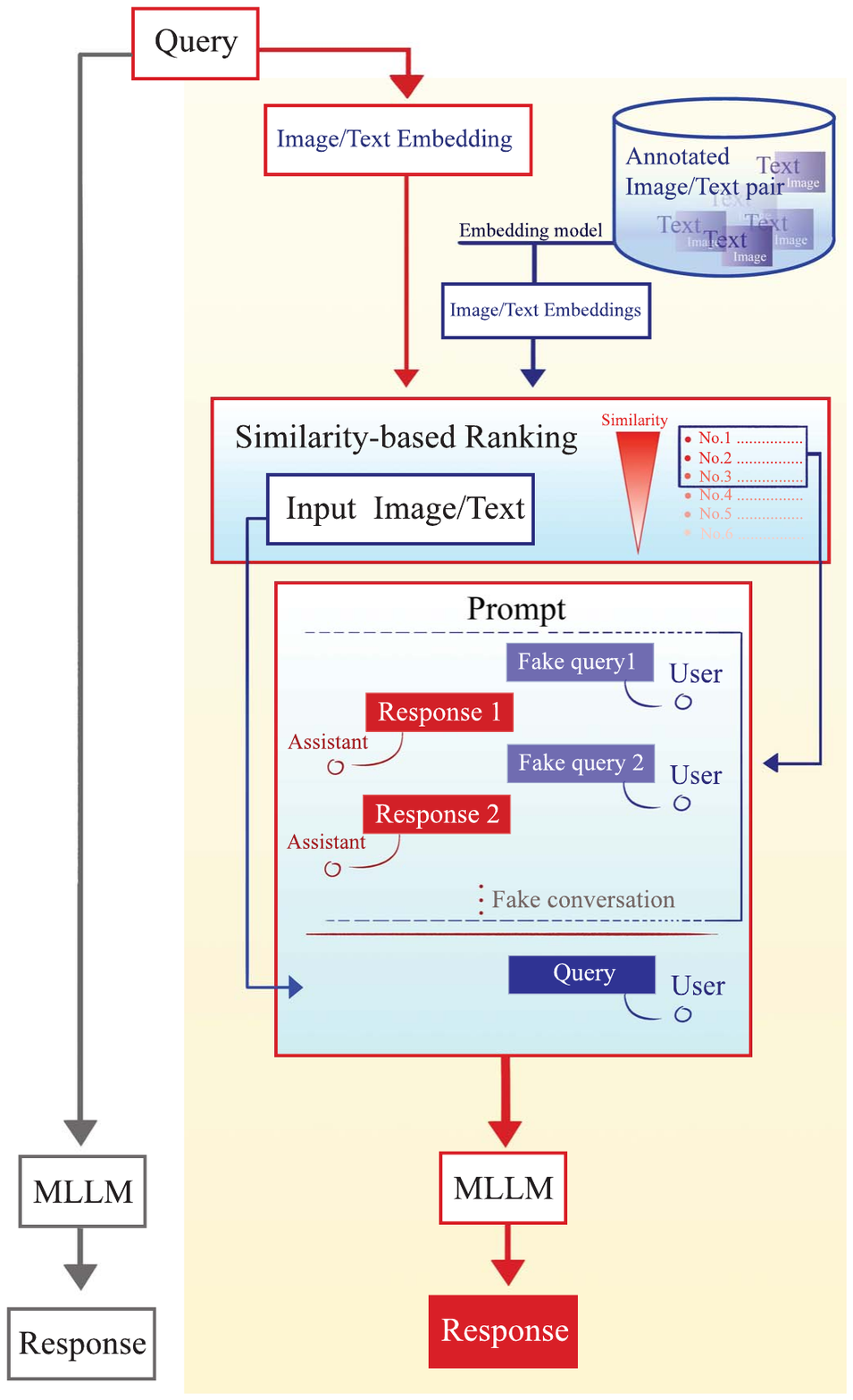}
    \caption{The overview of the retrieval-augmented in-context learning (RAICL) framework.}
    \label{fig:framework}
\end{figure}

\subsection{RAICL Framework}
The structure of the RAICL framework is illustrated in the figure \ref{fig:framework}. During inference, for each input query, all remaining image-pair data are treated as an external database. The input query's embedding is obtained either through an image-based embedding model or a text-based embedding model, while embeddings for all data points in the external database are generated using the same method. Similarity scores—primarily based on cosine similarity—are then computed between the query and the database entries to retrieve one or more of the most similar examples. These retrieved examples are incorporated into a carefully designed prompt, and the generated response serves as the final classification result. 
In this study, we tested our framework on Qwen2.5-VL-7B~\cite{bai2025qwen2}, Qwen2.5-VL-3B~\cite{bai2025qwen2}, LLaVA-v1.6-Mistral-7B~\cite{liu2024llavanext}, Gemma-3-4B~\cite{team2025gemma}, Gemma-3-12B~\cite{team2025gemma}.
Details regarding how embeddings are obtained, how similarity is computed, and how prompts are designed are provided in the following subsections.

\subsubsection{Embedding models}
To get the embeddings of images, we used ResNet-18/50/101~\cite{he2016deep} models. 
Each image is converted to three-channel RGB, resized to 224 × 224, and scaled to the range [0, 1]. The tensor is standardized with ImageNet~\cite{deng2009imagenet} statistics (mean [0.485, 0.456, 0.406], std [0.229, 0.224, 0.225]), given a batch dimension, and moved to GPU for inference. It is then passed through a ResNet-18/50/101~\cite{he2016deep} whose final fully connected layer has been replaced by an identity mapping; global average pooling yields a 512-dimensional feature that is finally L2-normalized onto the unit hypersphere for direct cosine-similarity computation.

For each corresponding text, the string is tokenized with a BERT~\cite{devlin2019bert} / BioBERT~\cite{lee2020biobert} / ClinicalBERT~\cite{alsentzer2019publicly} tokenizer, truncated or padded to 512 tokens, and fed to a frozen model on a GPU. The hidden state of the [CLS] token is extracted as a 768-dimensional sentence embedding. After gathering all non-empty embeddings, they are L2-normalized for further similarity computation.

\subsubsection{Similarity Metrics}
To illustrate the high compatibility of our framework, we employ several widely used similarity and distance metrics to measure the similarity between the input embedding and example embeddings.  
Specifically, we consider cosine similarity, inner product, Euclidean distance, Manhattan distance, and Chebyshev distance, each formally defined as follows.

Given two embedding vectors \( \mathbf{u}, \mathbf{v} \in \mathbb{R}^d \):

\begin{itemize}
    \item Cosine Similarity:  
    \[
    \text{cosine}(\mathbf{u}, \mathbf{v}) = \frac{\mathbf{u}^\top \mathbf{v}}{\|\mathbf{u}\|\|\mathbf{v}\|}
    \]
    % It measures the cosine of the angle between two vectors, focusing on their directional alignment.

    \item Inner Product:  
    \[
    \text{inner}(\mathbf{u}, \mathbf{v}) = \mathbf{u}^\top \mathbf{v}
    \]
    % It captures the raw interaction between vectors, sensitive to both magnitude and direction.

    \item Euclidean Distance:  
    \[
    \text{euclidean}(\mathbf{u}, \mathbf{v}) = \|\mathbf{u} - \mathbf{v}\|_2 = \sqrt{\sum_{i=1}^{d}(u_i - v_i)^2}
    \]
    % It computes the straight-line distance, reflecting overall proximity in the feature space.

    \item Manhattan Distance (also known as \( L_1 \) distance):  
    \[
    \text{manhattan}(\mathbf{u}, \mathbf{v}) = \|\mathbf{u} - \mathbf{v}\|_1 = \sum_{i=1}^{d} |u_i - v_i|
    \]
    % It sums absolute differences along each coordinate, robust to small axis-aligned changes.

    \item Chebyshev Distance:  
    \[
    \text{chebyshev}(\mathbf{u}, \mathbf{v}) = \|\mathbf{u} - \mathbf{v}\|_{\infty} = \max_{i=1,\dots,d} |u_i - v_i|
    \]
    % It measures the maximum single-axis deviation between vectors.
\end{itemize}

We used cosine similarity in most of our experiments and compared these similarity metrics in an ablation study.

\subsubsection{Prompt}
In our framework, we design a conversational-style prompt structure, as illustrated in Figure \ref{fig:framework}.  
Regardless of the number of examples selected, this prompt can be easily adapted in a consistent manner.  
At the bottom of the prompt, we place the image and text corresponding to the new query for disease classification.  
Above this, we simulate multiple rounds of dialogue based on the number of examples: if there is one example, we assume one prior round of conversation; if there are ten examples, we assume ten prior rounds.  
Each round of dialogue corresponds to one example, where the image and accompanying text are provided by the user side, embedded in a query.  
The assistant (i.e., the MLLM) then provides an accurate response. Importantly, the conversation is fake. We simulate the scenario where the MLLM has generated each response by itself, and the response is correct.
Through this design, all examples are embedded naturally into the prompt received by the MLLM, and the model only needs to imitate its previous answering style to respond to the new query.
We put the examples of the prompt and how to construct the prompt with multiple examples in Supplement Material 1. 

\subsubsection{Evaluation Metrics}

We evaluate model performance using a comprehensive set of metrics.  
Specifically, we report the overall accuracy (Acc), as well as micro-averaged precision, recall, and F1 score, which consider all instances equally and are sensitive to class imbalance.  
Additionally, we compute macro-averaged precision, recall, and F1 score by averaging metrics across classes without weighting, thus reflecting the model’s ability to perform uniformly across different categories.  
This combination of micro and macro metrics provides a balanced evaluation of both instance-level and class-level performance.

\subsubsection{Experimental Setup}
All experiments are conducted in an inference-only setting without any model fine-tuning.  
For experiments involving a single example, we perform inference on a single NVIDIA A100 GPU.  
For experiments involving multiple examples (especially for few-shot experiments), we distribute the inference workload across multiple NVIDIA A100 GPUs to accommodate the increased computational requirements.  
Our code is implemented using the Hugging Face Transformers library~\cite{wolf2020transformers}, leveraging its APIs for efficient model loading and generation.  
During model generation, we set the following parameters: temperature = 1.0, top\_k = 50, do\_sample = True, and num\_beams = 1.  
These settings ensure deterministic decoding while allowing the model to maintain a moderate level of response diversity when appropriate.

%% file: 3_results.tex
\section{Results}

\subsection{Main results}
Tables \ref{tab:diffmodeltcga} and \ref{tab:diffmodelschest} summarize a comprehensive study spanning five state-of-the-art multimodal language models—Qwen-2.5-VL-7B, Qwen-2.5-VL-3B, LLaVA-v1.6-Mistral-7B, Gemma-3-4B, and Gemma-3-12B. 
For the Qwen-2.5-VL-7B model on two datasets, comparing conditions with and without retrieval-augmented examples. The gains from RAG are evident on both benchmarks: baseline accuracies are below 0.80, whereas RAG raises them to 0.8368 on TCGA and 0.8658 on the IU Chest X-ray set. Every metric improves, and the Macro-F1 score on TCGA increases most dramatically—from 0.1191 to 0.8454—demonstrating the strong impact of retrieval-augmented in-context learning on multimodal large language models.

Table \ref{tab:diffmodeltcga} reports the results on the TCGA dataset. Examples selected with ResNet-18/50/101 image embeddings provide only modest gains, whereas those obtained with text encoders—BERT, BioBERT, or ClinicalBERT—yield substantial improvements across every model, implying that textual cues carry most of the discriminative signal in pathology reports.

Table \ref{tab:diffmodelschest} presents the corresponding evaluation on the IU Chest X-ray corpus. Here the pattern reverses: the largest gains appear when retrieval relies on ResNet-derived visual embeddings, while text-centric retrieval brings smaller benefits. These complementary trends indicate that the optimal retrieval strategy depends on the dataset’s dominant modality, yet in all cases the RAG framework enhances performance. 

Overall, introducing RAICL lifts accuracy by roughly ten percentage points (or more) over the non-RAG baselines, confirming the general effectiveness of retrieved in-context examples.

\subsection{Single modality vs. Multi-modality}
\label{sec_singlevsmultimodality}
We compared the performance of Qwen-2.5-VL-7B on the TCGA dataset under different input modalities. When the model receives only WSI images, accuracy is notably low (0.1701), highlighting the difficulty of diagnosing solely from slide visuals. Using text alone yields much higher classification accuracy (0.7764), indicating that the reports contain richer, more discriminative information. The best results arise (accuracy 0.7854) when both modalities are provided: integrating images with text further boosts performance, confirming that the two sources of evidence complement each other.

\begin{table}[htbp]
  \centering
  \renewcommand\arraystretch{0.8}
  \caption{One-shot performance comparison on TCGA datasets across representative MLLMs. For simplicity, we refer to the adopted encoders as the implemented RAICL.}
  \resizebox{\textwidth}{!}{
  \begin{tabular}{lccccccc}
    \toprule
    \multirow{2}{*}{Models}
      & \multirow{2}{*}{Acc}
      & \multicolumn{3}{c}{\textbf{Micro}}
      & \multicolumn{3}{c}{\textbf{Macro}} \\
    \cmidrule(lr){3-5}\cmidrule(lr){6-8}
      & & P & R & F1 & P & R & F1 \\
    \midrule
\multicolumn{8}{c}{\textbf{Qwen2.5-VL-7B}}\\
\midrule
Baseline (w/o RAG) & 0.7854 & 0.7854 & 0.7854 & 0.7854 & 0.1217 & 0.1245 & 0.1191\\
\hdashline
ResNet-18         & 0.7619 & 0.7619 & 0.7619 & 0.7619 & 0.6825 & 0.6955 & 0.6510\\
ResNet-50         & 0.7721 & 0.7721 & 0.7721 & 0.7721 & 0.4611 & 0.4687 & 0.4406\\
ResNet-101        & 0.7576 & 0.7576 & 0.7576 & 0.7576 & 0.5104 & 0.5178 & 0.4857\\
BERT              & 0.8192 & 0.8192 & 0.8192 & 0.8192 & 0.7119 & 0.7329 & 0.6955\\
ClinicalBERT      & 0.8240 & 0.8240 & 0.8240 & 0.8240 & \textbf{0.8580} & 0.8856 & 0.8397\\
BioBERT           & \textbf{0.8368} & \textbf{0.8368} & \textbf{0.8368} & \textbf{0.8368} & 0.8532 & \textbf{0.8862} & \textbf{0.8454}\\
    \midrule
    \multicolumn{8}{c}{\textbf{Qwen2.5-VL-3B}}\\
    \hdashline
    Baseline (w/o RAG) & 0.4002 & 0.4002 & 0.4002 & 0.4002 & 0.0292 & 0.0157 & 0.0172\\
    \hdashline
    ResNet-18         & 0.5581 & 0.5581 & 0.5581 & 0.5581 & 0.2245 & 0.2148 & 0.1811\\
    ResNet-50         & 0.5607 & 0.5607 & 0.5607 & 0.5607 & 0.2007 & 0.1914 & 0.1617\\
    ResNet-101        & \textbf{0.5650} &\textbf{ 0.5650} & \textbf{0.5650} & \textbf{0.5650} & 0.2143 & 0.2043 & 0.1731\\
    BERT              & 0.5425 & 0.5425 & 0.5425 & 0.5425 & 0.2293 & 0.2065 & 0.1750\\
    ClinicalBERT      & 0.5281 & 0.5281 & 0.5281 & 0.5281 & 0.2122 & 0.1875 & 0.1568\\
    BioBERT           & 0.5361 & 0.5361 & 0.5361 & 0.5361 & \textbf{0.2459} & \textbf{0.2178} & \textbf{0.1849}\\
    \midrule
    \multicolumn{8}{c}{\textbf{LLaVA-v1.6-Mistral-7B}}\\
    \hdashline
    Baseline (w/o RAG) & 0.7084 & 0.7084 & 0.7084 & 0.7084 & 0.0870 & 0.0785 & 0.0774\\
    \hdashline
    ResNet-18         & 0.7014 & 0.7014 & 0.7014 & 0.7014 & 0.1208 & 0.1086 & 0.1077\\
    ResNet-50         & 0.7057 & 0.7057 & 0.7057 & 0.7057 & 0.1256 & 0.1137 & 0.1131\\
    ResNet-101        & 0.6945 & 0.6945 & 0.6945 & 0.6945 & 0.1414 & 0.1271 & 0.1265\\
    BERT              & 0.7913 & 0.7913 & 0.7913 & 0.7913 & 0.1615 & 0.1533 & 0.1527\\
    ClinicalBERT      & \textbf{0.8159} & \textbf{0.8159} & \textbf{0.8159} & \textbf{0.8159} & \textbf{0.1715} & \textbf{0.1655} & \textbf{0.1643}\\
    BioBERT           & 0.8058 & 0.8058 & 0.8058 & 0.8058 & 0.1427 & 0.1369 & 0.1361\\
    \midrule
    \multicolumn{8}{c}{\textbf{Gemma-3-4B}}\\
    \hdashline
    Baseline (w/o RAG) & 0.4735 & 0.4735 & 0.4735 & 0.4735 & \textbf{0.1692} & \textbf{0.1688} & \textbf{0.1281}\\
    \hdashline
    ResNet-18         & 0.5078 & 0.5078 & 0.5078 & 0.5078 & 0.0921 & 0.0883 & 0.0689\\
    ResNet-50         & 0.4858 & 0.4858 & 0.4858 & 0.4858 & 0.0717 & 0.0664 & 0.0527\\
    ResNet-101        & 0.4912 & 0.4912 & 0.4912 & 0.4912 & 0.0665 & 0.0614 & 0.0484\\
    BERT              & 0.5024 & 0.5024 & 0.5024 & 0.5024 & 0.0970 & 0.0881 & 0.0710\\
    ClinicalBERT      & \textbf{0.5110} & \textbf{0.5110} & \textbf{0.5110} & \textbf{0.5110} & 0.0914 & 0.0824 & 0.0668\\
    BioBERT           & 0.5078 & 0.5078 & 0.5078 & 0.5078 & 0.0836 & 0.0752 & 0.0607\\
    \midrule
    \multicolumn{8}{c}{\textbf{Gemma-3-12B}}\\
    \hdashline
    Baseline (w/o RAG) & 0.6854 & 0.6854 & 0.6854 & 0.6854 & 0.3214 & 0.3230 & 0.2942\\
    \hdashline
    ResNet-18         & 0.8919 & 0.8919 & 0.8919 & 0.8919 & 0.7214 & 0.7363 & 0.7244\\
    ResNet-50         & 0.8914 & 0.8914 & 0.8914 & 0.8914 & 0.8650 & 0.8818 & 0.8680\\
    ResNet-101        & 0.8807 & 0.8807 & 0.8807 & 0.8807 & 0.8538 & 0.8743 & 0.8576\\
    BERT              & \textbf{0.9192} & \textbf{0.9192} & \textbf{0.9192} & \textbf{0.9192} & \textbf{0.8950} &\textbf{ 0.8979} & \textbf{0.8914}\\
    ClinicalBERT      & 0.9171 & 0.9171 & 0.9171 & 0.9171 & 0.8932 & 0.8917 & 0.8857\\
    BioBERT           & 0.9171 & 0.9171 & 0.9171 & 0.9171 & 0.8938 & 0.8928 & 0.8866\\
    \bottomrule
  \end{tabular}}
\label{tab:diffmodeltcga}
\end{table}

% \subsection{Robustness on different models}
% Tables \ref{tab:diffmodeltcga} and \ref{tab:diffmodelschest} summarize a robustness study spanning four state-of-the-art multimodal language models—Qwen-2.5-VL-3B, LLaVA-v1.6-Mistral-7B, Gemma-3-4B, and Gemma-3-12B. Introducing RAG lifts accuracy by roughly ten percentage points (or more) over the non-RAG baselines, confirming the general effectiveness of retrieved in-context examples.

% Table \ref{tab:diffmodeltcga} reports the results on the TCGA dataset. Examples selected with ResNet-18/50/101 image embeddings provide only modest gains, whereas those obtained with text encoders—BERT, BioBERT, or ClinicalBERT—yield substantial improvements across every model, implying that textual cues carry most of the discriminative signal in pathology reports.

% Table \ref{tab:diffmodelschest} presents the corresponding evaluation on the IU Chest X-ray corpus. Here the pattern reverses: the largest gains appear when retrieval relies on ResNet-derived visual embeddings, while text-centric retrieval brings smaller benefits. These complementary trends indicate that the optimal retrieval strategy depends on the dataset’s dominant modality, yet in all cases the RAG framework enhances performance.

\begin{table}[htbp]
  \centering
  \renewcommand\arraystretch{0.8}
  \caption{One-shot performance comparison on IU Chest X-ray dataset across different state-of-the-art MLLMs. For simplicity, we refer to the adopted encoders as the implemented RAICL.}
  \label{tab:diffmodelschest}
  \resizebox{\textwidth}{!}{
  \begin{tabular}{lccccccc}
    \toprule
    \multirow{2}{*}{Models}
      & \multirow{2}{*}{Acc}
      & \multicolumn{3}{c}{\textbf{Micro}}
      & \multicolumn{3}{c}{\textbf{Macro}} \\
    \cmidrule(lr){3-5}\cmidrule(lr){6-8}
      & & P & R & F1 & P & R & F1 \\
\midrule
\multicolumn{8}{c}{\textbf{Qwen2.5-VL-7B}}\\
\midrule
Baseline (w/o RAG) & 0.7924 & 0.7924 & 0.7924 & 0.7924 & 0.3780 & 0.3851 & 0.3564\\
\hdashline
ResNet-18         & \textbf{0.8658} & \textbf{0.8658} & \textbf{0.8658} & \textbf{0.8658} & \textbf{0.5027} & 0.4003 & 0.3981\\
ResNet-50         & 0.8645 & 0.8645 & 0.8645 & 0.8645 & 0.4640 & 0.3906 & 0.3912\\
ResNet-101        & 0.8578 & 0.8578 & 0.8578 & 0.8578 & 0.4384 & 0.3782 & 0.3773\\
BERT              & 0.8591 & 0.8591 & 0.8591 & 0.8591 & 0.4891 & \textbf{0.4211} & 0.4111\\
ClinicalBERT      & 0.8551 & 0.8551 & 0.8551 & 0.8551 & 0.4710 & 0.4085 & 0.4000\\
BioBERT           & 0.8565 & 0.8565 & 0.8565 & 0.8565 & 0.4734 & 0.4174 & \textbf{0.4164}\\
    \midrule
\multicolumn{8}{c}{\textbf{Qwen2.5-VL-3B}} \\
\hdashline
Baseline (w/o RAG) & 0.8164 & 0.8164 & 0.8164 & 0.8164 & 0.4242 & 0.4557 & 0.4199 \\
\hdashline
Resnet18   & \textbf{0.9099} & \textbf{0.9099} & \textbf{0.9099} & \textbf{0.9099} & \textbf{0.5466} & \textbf{0.6247} & \textbf{0.5687} \\
Resnet50   & 0.9092 & 0.9092 & 0.9092 & 0.9092 & 0.4701 & 0.5199 & 0.4771 \\
Resnet101  & 0.9005 & 0.9005 & 0.9005 & 0.9005 & 0.5266 & 0.6008 & 0.5441 \\
BERT       & 0.8618 & 0.8618 & 0.8618 & 0.8618 & 0.4385 & 0.4987 & 0.4289 \\
ClinicalBERT & 0.8685 & 0.8685 & 0.8685 & 0.8685 & 0.4321 & 0.5073 & 0.4327 \\
BioBERT    & 0.8665 & 0.8665 & 0.8665 & 0.8665 & 0.4807 & 0.5208 & 0.4509 \\
\midrule
\multicolumn{8}{c}{\textbf{llava-v1.6-mistral-7b}} \\
\hdashline
Baseline (w/o RAG) & 0.8785 & 0.8785 & 0.8785 & 0.8785 & 0.2233 & 0.2210 & 0.2069 \\
\hdashline
Resnet18   & 0.9052 & 0.9052 & 0.9052 & 0.9052 & 0.2341 & 0.2121 & 0.2139 \\
Resnet50   & 0.9032 & 0.9032 & 0.9032 & 0.9032 & 0.2194 & 0.1975 & 0.1958 \\
Resnet101  & 0.9012 & 0.9012 & 0.9012 & 0.9012 & 0.2228 & 0.1927 & 0.1943 \\
BERT       & 0.9206 & 0.9206 & 0.9206 & 0.9206 & 0.2296 & 0.2165 & 0.2143 \\
ClinicalBERT & 0.9186 & 0.9186 & 0.9186 & 0.9186 & 0.2706 & \textbf{0.2505} & 0.2474 \\
BioBERT    & \textbf{0.9332} & \textbf{0.9332} & \textbf{0.9332} & \textbf{0.9332} & \textbf{0.2811} & 0.2488 & \textbf{0.2542} \\
\midrule
\multicolumn{8}{c}{\textbf{Gemma-3-4b}} \\
\hdashline
Baseline (w/o RAG) & 0.2570 & 0.2570 & 0.2570 & 0.2570 & 0.3270 & 0.3549 & 0.2900 \\
\hdashline
Resnet18   & 0.7503 & 0.7503 & 0.7503 & 0.7503 & \textbf{0.5674} & \textbf{0.6720} & \textbf{0.5589} \\
Resnet50   & 0.7470 & 0.7470 & 0.7470 & 0.7470 & 0.5308 & 0.6543 & 0.5439 \\
Resnet101  & \textbf{0.7623} & \textbf{0.7623} & \textbf{0.7623} & \textbf{0.7623} & 0.4437 & 0.5407 & 0.4455 \\
BERT       & 0.4112 & 0.4112 & 0.4112 & 0.4112 & 0.3705 & 0.4299 & 0.3447 \\
ClinicalBERT & 0.4246 & 0.4246 & 0.4246 & 0.4246 & 0.3972 & 0.4582 & 0.3707 \\
BioBERT    & 0.4166 & 0.4166 & 0.4166 & 0.4166 & 0.4134 & 0.4844 & 0.3869 \\
\midrule
\multicolumn{8}{c}{\textbf{Gemma-3-12b}} \\
\hdashline
Baseline (w/o RAG) & 0.8812 & 0.8812 & 0.8812 & 0.8812 & 0.5627 & \textbf{0.8206} & 0.6417 \\
\hdashline
Resnet18   & \textbf{0.9112} & \textbf{0.9112} & \textbf{0.9112} & \textbf{0.9112} & 0.6272 & 0.7539 & 0.6506 \\
Resnet50   & 0.9039 & 0.9039 & 0.9039 & 0.9039 & 0.6029 & 0.7636 & 0.6488 \\
Resnet101  & 0.9072 & 0.9072 & 0.9072 & 0.9072 & \textbf{0.6342} & 0.7612 & \textbf{0.6583} \\
BERT       & 0.8652 & 0.8652 & 0.8652 & 0.8652 & 0.5657 & 0.7618 & 0.6080 \\
ClinicalBERT & 0.8705 & 0.8705 & 0.8705 & 0.8705 & 0.5704 & 0.7686 & 0.6214 \\
BioBERT    & 0.8632 & 0.8632 & 0.8632 & 0.8632 & 0.5620 & 0.7656 & 0.6129 \\
    \bottomrule
  \end{tabular}}
\end{table}

% \begin{table}[htbp]\centering
% % \renewcommand\arraystretch{1.1}
% \caption{Accuracy comparison between single modality and multi-modality using Qwen2.5-VL-7B model on TCGA dataset.}
% \begin{tabular}{lcccccccc}
% \toprule
% Modality & Acc\\
% \midrule
% Image only & 0.1701 \\
% Text only  & 0.7764 \\
% Multi-modality & 0.7854 \\
% \bottomrule
% \end{tabular}
% \label{tab:singlemodality}
% \end{table}
% % Image only & 0.1701 & 0.1701 & 0.1701 & 0.1701 & 0.2119 & 0.2047 & 0.0833\\
% % Text only  & 0.7764 & 0.7764 & 0.7764 & 0.7764 & 0.1484 & 0.1553 & 0.1463\\
% % Mulit-modality & 0.7854 & 0.7854 & 0.7854 & 0.7854 & 0.1217 & 0.1245 & 0.1191\\

\subsection{Robustness on few-shot settings}
To show that the framework is naturally suited to in-context learning, we compared its performance as the number of retrieved exemplars (k-shot) increases. Table \ref{tab:fewshottcga} reports Qwen-2.5-VL-7B results on the TCGA dataset when examples are retrieved with cosine similarity derived from various embedding models. With the sole exception of the ResNet-50 encoder, every setting exhibits a clear upward trend: more examples consistently translate into higher accuracy.
Table \ref{tab:fewshotchest} gives the corresponding analysis on the IU Chest X-ray corpus. When text-based encoders are used for retrieval, the same monotonic improvement is observed. In contrast, image-based encoders peak at 5-shot, with 10-shot offering no further benefit, yielding virtually identical scores.

\begin{table}[htbp]
\caption{Performance of Qwen2.5-VL-7B model for different few-shot settings on the TCGA dataset.}
\centering
\renewcommand\arraystretch{0.8}
\resizebox{\textwidth}{!}{
\begin{tabular}{ccccccccc}
\toprule
\multirow{2}{*}{Encoder} & \multirow{2}{*}{Setting} & \multirow{2}{*}{Acc} &
\multicolumn{3}{c}{Micro} & \multicolumn{3}{c}{Macro}\\
\cmidrule(lr){4-6}\cmidrule(lr){7-9}
 & & & P & R & F1 & P & R & F1\\
\midrule
\multirow{3}{*}{ResNet-18}
 & 1 shot  & 0.7619 & 0.7619 & 0.7619 & 0.7619 & 0.6825 & 0.6955 & 0.6510\\
 & 5 shot  & 0.7555 & 0.7555 & 0.7555 & 0.7555 & 0.8531 & 0.8493 & 0.7891\\
 & 10 shot & \textbf{0.7774} & \textbf{0.7774} & \textbf{0.7774} & \textbf{0.7774} & \textbf{0.8536} & \textbf{0.8585} & \textbf{0.8058}\\
\midrule
\multirow{3}{*}{ResNet-50}
 & 1 shot  & \textbf{0.7721} & \textbf{0.7721} & \textbf{0.7721} & \textbf{0.7721} & 0.4611 & 0.4687 & 0.4406\\
 & 5 shot  & 0.7464 & 0.7464 & 0.7464 & 0.7464 & 0.7088 & 0.7027 & 0.6511\\
 & 10 shot & 0.7587 & 0.7587 & 0.7587 & 0.7587 & \textbf{0.7111} & \textbf{0.7083} & \textbf{0.6599}\\
\midrule
\multirow{3}{*}{ResNet-101}
 & 1 shot  & 0.7576 & 0.7576 & 0.7576 & 0.7576 & 0.5104 & 0.5178 & 0.4857\\
 & 5 shot  & 0.7528 & 0.7528 & 0.7528 & 0.7528 & 0.7096 & 0.7047 & 0.6556\\
 & 10 shot & \textbf{0.7710} & \textbf{0.7710} & \textbf{0.7710} & \textbf{0.7710} & \textbf{0.8551} & \textbf{0.8565} & \textbf{0.8016}\\
\midrule
\multirow{3}{*}{BERT}
 & 1 shot  & 0.8192 & 0.8192 & 0.8192 & 0.8192 & 0.7119 & 0.7329 & 0.6955\\
 & 5 shot  & 0.8566 & 0.8566 & 0.8566 & 0.8566 & 0.8798 & 0.9084 & 0.8690\\
 & 10 shot & \textbf{0.8700} & \textbf{0.8700} & \textbf{0.8700} & \textbf{0.8700} & \textbf{0.8863} & \textbf{0.9174} & \textbf{0.8800}\\
\midrule
\multirow{3}{*}{ClinicalBERT}
 & 1 shot  & 0.8240 & 0.8240 & 0.8240 & 0.8240 & 0.8580 & 0.8856 & 0.8397\\
 & 5 shot  & 0.8561 & 0.8561 & 0.8561 & 0.8561 & 0.7347 & 0.7559 & 0.7242\\
 & 10 shot & \textbf{0.8764} & \textbf{0.8764} & \textbf{0.8764} & \textbf{0.8764} & \textbf{0.8921} & \textbf{0.9219} & \textbf{0.8861}\\
\midrule
\multirow{3}{*}{BioBERT}
 & 1 shot  & 0.8368 & 0.8368 & 0.8368 & 0.8368 & 0.8532 & 0.8862 & 0.8454\\
 & 5 shot  & 0.8513 & 0.8513 & 0.8513 & 0.8513 & 0.8830 & 0.9082 & 0.8670\\
 & 10 shot & \textbf{0.8646} & \textbf{0.8646} & \textbf{0.8646} & \textbf{0.8646} & \textbf{0.8865} & \textbf{0.9133} & \textbf{0.8762}\\
\bottomrule
\end{tabular}}
\label{tab:fewshottcga}
\end{table}

\begin{table}[htbp]
\caption{Performance of Qwen2.5-VL-7B model for different few-shot settings on the IU Chest X-ray dataset. }
\centering
\renewcommand\arraystretch{0.8}   % 行距微调
\resizebox{\textwidth}{!}{
\begin{tabular}{ccccccccc}
\toprule
\multirow{2}{*}{Encoder} & \multirow{2}{*}{Setting} & \multirow{2}{*}{Acc} &
\multicolumn{3}{c}{Micro} & \multicolumn{3}{c}{Macro}\\
\cmidrule(lr){4-6}\cmidrule(lr){7-9}
 & & & P & R & F1 & P & R & F1\\
\midrule
\multirow{3}{*}{ResNet-18} 
  & 1 shot  & 0.8658          & 0.8658          & 0.8658          & 0.8658          & 0.5027          & 0.4003          & 0.3981\\
  & 5 shot  & \textbf{0.9172} & \textbf{0.9172} & \textbf{0.9172} & \textbf{0.9172} & \textbf{0.6849} & \textbf{0.5889} & 0.5774\\
  & 10 shot & 0.9166          & 0.9166          & 0.9166          & 0.9166          & 0.6822          & \textbf{0.5889} & \textbf{0.5862}\\
\midrule
\multirow{3}{*}{ResNet-50} 
  & 1 shot  & 0.8645          & 0.8645          & 0.8645          & 0.8645          & 0.4640          & 0.3906          & 0.3912\\
  & 5 shot  & \textbf{0.9172} & \textbf{0.9172} & \textbf{0.9172} & \textbf{0.9172} & 0.6324          & 0.5970          & 0.5882\\
  & 10 shot & 0.9166          & 0.9166          & 0.9166          & 0.9166          & \textbf{0.6553} & \textbf{0.6040} & \textbf{0.5942}\\
\midrule
\multirow{3}{*}{ResNet-101} 
  & 1 shot  & 0.8578          & 0.8578          & 0.8578          & 0.8578          & 0.4384          & 0.3782          & 0.3773\\
  & 5 shot  & \textbf{0.9172} & \textbf{0.9172} & \textbf{0.9172} & \textbf{0.9172} & \textbf{0.6983} & 0.5795          & \textbf{0.5808}\\
  & 10 shot & 0.9152          & 0.9152          & 0.9152          & 0.9152          & 0.6502          & \textbf{0.5804} & 0.5777\\
\midrule
\multirow{3}{*}{BERT} 
  & 1 shot  & 0.8591          & 0.8591          & 0.8591          & 0.8591          & 0.4891          & 0.4211          & 0.4111\\
  & 5 shot  & 0.9252          & 0.9252          & 0.9252          & 0.9252          & 0.7292          & 0.6709          & 0.6590\\
  & 10 shot & \textbf{0.9299} & \textbf{0.9299} & \textbf{0.9299} & \textbf{0.9299} & \textbf{0.7191} & \textbf{0.7006} & \textbf{0.6746}\\
\midrule
\multirow{3}{*}{ClinicalBERT} 
  & 1 shot  & 0.8565          & 0.8565          & 0.8565          & 0.8565          & 0.4734          & 0.4174          & 0.4164\\
  & 5 shot  & 0.9319          & 0.9319          & 0.9319          & 0.9319          & 0.7024          & 0.6763          & 0.6577\\
  & 10 shot & \textbf{0.9326} & \textbf{0.9326} & \textbf{0.9326} & \textbf{0.9326} & \textbf{0.7363} & \textbf{0.7066} & \textbf{0.6922}\\
\midrule
\multirow{3}{*}{BioBERT} 
  & 1 shot  & 0.8551          & 0.8551          & 0.8551          & 0.8551          & 0.4710          & 0.4085          & 0.4000\\
  & 5 shot  & 0.9319          & 0.9319          & 0.9319          & 0.9319          & 0.7214          & 0.7072          & 0.6813\\
  & 10 shot & \textbf{0.9352} & \textbf{0.9352} & \textbf{0.9352} & \textbf{0.9352} & \textbf{0.7465} & \textbf{0.7176} & \textbf{0.6991}\\
\bottomrule
\end{tabular}
}
\label{tab:fewshotchest}
\end{table}

\begin{table}[ht]
\centering
\caption{One-shot performance comparison of different similarity metrics on the TCGA dataset.}
\resizebox{\textwidth}{!}{
\begin{tabular}{llcccccccc}
\toprule
\multirow{2}{*}{Encoder} & \multirow{2}{*}{Similarity metrics} & \multirow{2}{*}{Acc} & \multicolumn{3}{c}{Micro} & \multicolumn{3}{c}{Macro} \\
\cmidrule(lr){4-6} \cmidrule(lr){7-9}
 & & & p & r & f1 & p & r & f1 \\
\midrule
\multirow{5}{*}{Resnet18} & Cosine        & 0.7619 & 0.7619 & 0.7619 & 0.7619 & \textbf{0.6825} & \textbf{0.6955} & \textbf{0.651} \\
      & Inner product & 0.7416 & 0.7416 & 0.7416 & 0.7416 & 0.5036 & 0.5097 & 0.476 \\
      & Euclidean     & \textbf{0.7774} & \textbf{0.7774} & \textbf{0.7774} & \textbf{0.7774} & 0.5178 & 0.5278 & 0.4976\\
      & Manhattan     & 0.7726 & 0.7726 & 0.7726 & 0.7726 & 0.5175 & 0.5266 & 0.4963 \\
      & Chebyshev     & 0.7598 & 0.7598 & 0.7598 & 0.7598 & 0.5123 & 0.5198 & 0.4876 \\
\midrule
\multirow{5}{*}{BERT}     & Cosine        & 0.8192 & 0.8192 & 0.8192 & 0.8192 & \textbf{0.7119} & \textbf{0.7329} & \textbf{0.6955} \\
      & Inner product & 0.7742 & 0.7742 & 0.7742 & 0.7742 & 0.5968 & 0.6081 & 0.569 \\
      & Euclidean     & \textbf{0.8240} & \textbf{0.8240} & \textbf{0.8240} & \textbf{0.8240} & 0.4227 & 0.4358 & 0.4167 \\
      & Manhattan     & 0.8224 & 0.8224 & 0.8224 & 0.8224 & 0.4226 & 0.4351 & 0.4361 \\
      & Chebyshev     & 0.8095 & 0.8095 & 0.8095 & 0.8095 & 0.4640 & 0.4773 & 0.4554 \\
\bottomrule
\end{tabular}
}
\label{tab:simcomp}
\end{table}

\subsection{Robustness on different metrics}

Table \ref{tab:simcomp} explores the effect of using different similarity metrics. After producing embeddings with either ResNet-18 or BERT, we retrieve examples with cosine similarity, inner product, Euclidean, Manhattan, and Chebyshev distances and then evaluate on the TCGA dataset. Euclidean distance delivers the highest accuracy, whereas cosine similarity yields the best macro-F1 score, suggesting that the optimal metric depends on which evaluation criterion is prioritized.

%% file: 4_Discussion.tex
\section{Discussion}
\label{sec_dis}

This study aimed to improve disease classification by utilizing a novel Retrieval-Augmented In-Context Learning (RAICL) framework.
The RAICL framework leverages two advanced techniques—RAG and ICL—to significantly enhance the capabilities of MLLMs in disease classification. One of the most notable outcomes of our study is the effectiveness of using retrieval-augmented examples, which enriched the context provided to the model. This results in a much more robust and accurate disease classification system. For example, the introduction of RAG led to dramatic improvements in the performance of the Qwen-2.5-VL-7B model across various embedding models, such as ResNet and BERT. As demonstrated in Table \ref{tab:diffmodeltcga} and \ref{tab:diffmodelschest}, the integration of medical images with clinical text consistently provided the best results, surpassing the performance of models that relied on images alone.

% \subsection{The Importance of Multi-modality}
The comparison between single-modality and multi-modality models is a crucial aspect of this study, as mentioned in subsection \ref{sec_singlevsmultimodality}, 
The integration of both images and text resulted in better performance than using either modality alone. This suggests that each modality—images and text—contains complementary information, which when combined, leads to a more holistic understanding of the disease classification problem. Specifically, combining histopathology images with corresponding pathology reports improved accuracy from 0.1701 (image only) to 0.7854 (multi-modality), highlighting the rich contextual information that text provides in the classification task.

% \subsection{Comparison with State-of-the-art Models}
The comparison of RAICL with other state-of-the-art multimodal models, such as Qwen-2.5-VL-3B, LLaVA-v1.6-Mistral-7B, and Gemma-3 models, reveals several important insights. As seen in Tables \ref{tab:diffmodeltcga} and \ref{tab:diffmodelschest}, the RAICL framework consistently outperformed the baseline models across all datasets. The improvements were especially noticeable when retrieval relied on text-based models (e.g., BERT, BioBERT, ClinicalBERT), which provided substantial performance boosts on the TCGA dataset. This highlights the significant role that clinical reports play in disease diagnosis, especially for pathology and radiology-based tasks. On the IU Chest X-ray dataset, the largest gains were observed when using visual embeddings (e.g., ResNet-based retrieval), suggesting that the optimal modality for retrieval depends on the nature of the dataset.
For each dataset, the information contained in images and text is different, and the richness of information in each modality is directly related to classification performance. If the information in the text is more abundant and helps with classification, using the text's embedding to retrieve examples would be a better choice. Conversely, if the information in the image is more valuable, retrieving similar examples using the image's embedding will yield better results.
This reinforces the idea that RAICL's flexibility in combining multiple modalities enhances its robustness and performance across different tasks.

% \subsection{Analysis of Few-shot Learning Performance}
One of the most compelling findings of this study is the ability of the RAICL framework to perform well in few-shot learning settings. As shown in Tables \ref{tab:fewshottcga} and \ref{tab:fewshotchest}, increasing the number of retrieved exemplars led to consistent improvements in model accuracy. This is particularly important in real-world medical scenarios where annotated data is often limited. The RAICL framework's ability to generalize well with minimal data makes it highly valuable for medical diagnosis tasks that involve rare or underrepresented diseases. Furthermore, the improved accuracy with text-based retrieval methods suggests that textual information plays a crucial role in guiding the model toward accurate classifications.

% \subsection{Robustness Across Models and Tasks}
The results from various state-of-the-art MLLMs, including the Qwen-2.5-VL-7B, LLaVA-v1.6-Mistral-7B, and Gemma-3 models, further emphasize the robustness of the RAICL framework. Our experiments show that RAICL consistently outperforms the baseline models by approximately 10 percentage points in accuracy, regardless of the underlying model architecture. This is particularly important in the context of biomedical applications, where high accuracy is critical for clinical decision-making. The framework's ability to enhance performance across different models and tasks (e.g., TCGA and IU Chest X-ray datasets) confirms its versatility and potential for widespread use in medical diagnosis.

% \subsection{Limitations of the RAICL Framework}
While the RAICL framework shows promising results, there are several limitations to consider. First, the computational cost of using large multimodal datasets is non-trivial, especially when employing large models such as those with 70B or more parameters, which require substantial computational resources for inference. Secondly, this study only explored two modalities—image and text—whereas real-world applications often involve a wider range of modalities. Therefore, further research is needed to extend RAICL’s capabilities to more complex, multimodal scenarios encountered in practice. 

% \subsection{Future Work}
Several avenues for future research can further enhance the RAICL framework's performance and applicability. First, exploring the use of additional similarity metrics, such as Manhattan and Euclidean distances, could help refine the retrieval process and improve classification accuracy (as shown in Table \ref{tab:simcomp}). Second, we plan to investigate ways to optimize the framework's efficiency, particularly in terms of reducing computational costs without sacrificing accuracy~\cite{zhan2025epee}. Quantization methods~\cite{jin2024efficient,li2024large} and multi-task learning~\cite{zhan2024MTL,zhan2025ramie} could offer a promising direction to make RAICL more feasible for deployment in resource-constrained environments. Finally, future studies should explore the application of RAICL to a broader range of modalities, such as video, speech, and time series signals.

%% file: 5_conclusion.tex
\section{Conclusion}
\label{sec_con}

In this study, we proposed the RAICL framework to enhance MLLMs in disease classification. By integrating RAG with ICL, RAICL effectively leverages both visual and textual modalities to provide richer context for diagnosis.
Extensive experiments on two real-world datasets, the TCGA dataset and the IU Chest X-ray dataset, demonstrated that RAICL consistently improves classification performance across a range of state-of-the-art MLLMs. The results highlight that retrieval-augmented examples substantially enhance model robustness and accuracy, with text-based retrieval proving particularly effective for pathology tasks and image-based retrieval for radiology tasks. Furthermore, RAICL shows strong generalization in few-shot settings, indicating its potential for efficient deployment in clinical applications where labeled data is limited.
Overall, our findings establish RAICL as a powerful and versatile framework for advancing multimodal disease classification, offering significant improvements without requiring additional model fine-tuning.

\section{CRediT authorship contribution statement}
\textbf{Zaifu Zhan}: 
Conceptualization, Data curation, Formal analysis, Investigation, Methodology, Validation, Writing – original draft, Writing – review and editing. 
\textbf{Shuang Zhou}:
Conceptualization, Data curation, Methodology, Validation, Writing – review and editing
\textbf{Xiaoshan Zhou}:
Writing – review and editing.
\textbf{Yongkang Xiao}:
Writing – review and editing.
\textbf{Jun Wang}:
Writing – review and editing.
\textbf{Jiawen Deng}:
Writing – review and editing.
\textbf{He Zhu}:
Writing – review and editing.
\textbf{Yu Hou}:
Writing – review and editing.
\textbf{Rui Zhang}:
Conceptualization, Funding acquisition, Project administration, Resources, Supervision, Writing – review and editing

\section{Declaration of competing interest}
The authors declare that they have no known competing financial interests or personal relationships that could have appeared to influence the work reported in this paper.

\section{Data availability}
Readers could find the datasets we used through the following links:

TCGA:\href{https://www.cancer.gov/ccg/research/genome-sequencing/tcga}{https://www.cancer.gov/ccg/research/genome-sequencing/tcga}

IU Chest X-ray: \href{https://www.kaggle.com/datasets/raddar/chest-xrays-indiana-university}{https://www.kaggle.com/datasets/raddar/chest-xrays-indiana-university}

We will release the code upon the acceptance of this paper.

\section{Acknowledgements}
This work was supported by the National Center for Complementary and Integrative Health [grant numbers R01AT009457, U01AT012871]; the National Institute on Aging [grant number R01AG078154]; the National Cancer Institute [grant number R01CA287413]; the National Institute of Diabetes and Digestive and Kidney Diseases [grant number R01DK115629]; and the National Institute on Minority Health and Health Disparities [grant number 1R21MD019134-01].

\section{Declaration of generative AI and AI-assisted technologies in the writing process}
During the preparation of this work, the authors used ChatGPT to check grammar. After using this tool, the authors reviewed and edited the content as needed and take full responsibility for the content of the published article.